\documentclass[runningheads]{llncs}

\usepackage{eccv}

\usepackage{eccvabbrv}

\usepackage{graphicx}
\usepackage{booktabs}
\usepackage{multirow}
\usepackage{array}

\usepackage[normalem]{ulem}
\useunder{\uline}{\ul}{}

\usepackage[accsupp]{axessibility}  

\usepackage{hyperref}

\usepackage{orcidlink}

\begin{document}

\title{Dual-Image Enhanced CLIP for Zero-Shot Anomaly Detection} 



\author{Zhaoxiang Zhang\inst{1} \and
Hanqiu Deng\inst{1} \and
Jinan Bao\inst{1} \and
Xingyu Li\inst{1}}

\authorrunning{Zhang et al.}

\institute{University of Alberta, Edmonton AB T6J2R3, Canada}

\maketitle

\begin{abstract}
  Image Anomaly Detection has been a challenging task in Computer Vision field. The advent of Vision-Language models, particularly the rise of CLIP-based frameworks, has opened new avenues for zero-shot anomaly detection. Recent studies have explored the use of CLIP by aligning images with normal and prompt descriptions. However, the exclusive dependence on textual guidance often falls short, highlighting the critical importance of additional visual references. In this work, we introduce a Dual-Image Enhanced CLIP approach, leveraging a joint vision-language scoring system. Our methods process pairs of images, utilizing each as a visual reference for the other, thereby enriching the inference process with visual context. This dual-image strategy markedly enhanced both anomaly classification and localization performances. Furthermore, we have strengthened our model with a test-time adaptation module that incorporates synthesized anomalies to refine localization capabilities. Our approach significantly exploits the potential of vision-language joint anomaly detection and demonstrates comparable performance with current SOTA methods across various datasets.
\end{abstract}

\section{Introduction}
\label{sec:intro}

\begin{figure}
    \centering
    \includegraphics[width=0.95\textwidth]{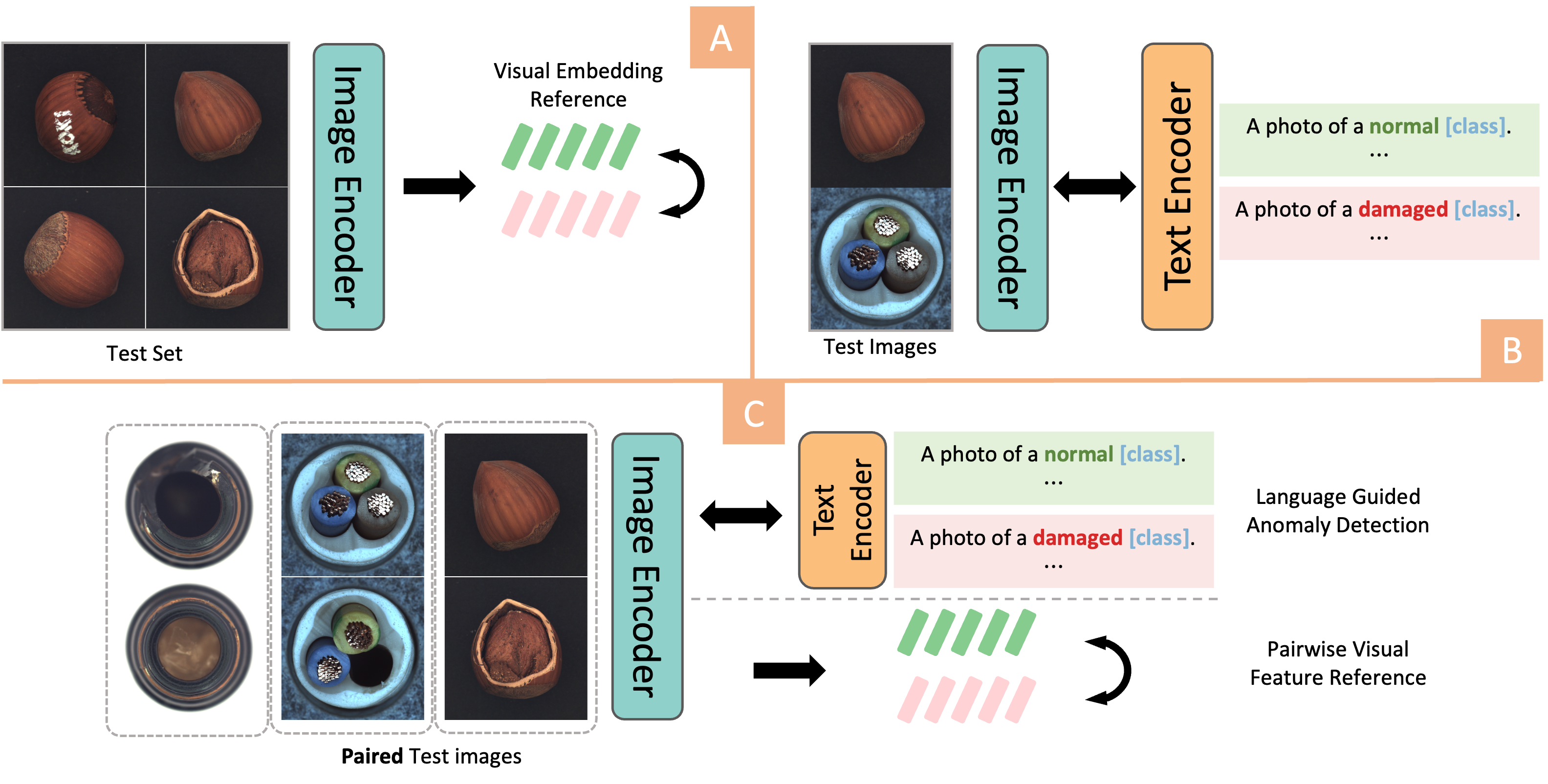}
    \caption{Overview of the Dual-Image Enhanced CLIP Anomaly Detection Model. Traditional approaches often depend on a single modality for anomaly detection, where (A) demonstrates the use of image embeddings, and (B) illustrates reliance on text prompts. Our proposed method, shown in (C), integrates both visual and textual information, utilizing a dual-image input to enrich the feature space for a more robust and comprehensive anomaly detection framework.}
    \label{fig: overview}
\end{figure}

Visual anomaly detection (AD) and localization are crucial aspects of computer vision, finding widespread applications in industrial inspection \cite{mvtec, visa}, medical imaging \cite{fanogan}, and video surveillance \cite{interpolatin_frame}. This complex task involves identifying and pinpointing atypical patterns, deviations, or anomalies in visual data. These are often characterized by subtle differences in texture, color, shape, or motion, blending seamlessly into normal surroundings. Due to their diverse nature, AD poses significant challenges and has been the focus of extensive research in real-world applications.

Historically, anomaly detection has been approached through one-class methods, training unique models for each category of normal images \cite{padim, patchcore, mkd, rd, fanogan, draem}. These methods achieve high accuracy on public benchmarks \cite{mvtec, visa} when abundant normal images are available. However, in the scenario where few or no normal training data are available, these approaches may not be ideal. Zero-shot anomaly detection (ZSAD) emerges as a vital task in such scenarios, requiring models to detect anomalies without training samples from a target dataset. This study focuses on ZSAD scenario.

MuSc \cite{musc} was proposed to leverage unlabelled images in the test set as references for the query images. It operates under the argument that a rich amount of normal information implicit in unlabelled test images is underutilized. Even if the test image is anomalous, it still contains some normal patches that can serve as references. MuSc achieves SOTA performance, but as it requires knowledge of the test set distributions before inference, it aligns more with transductive rather than inductive learning. Also, its extensive comparison with all test set images can be time-consuming and computationally intensive.

Alternatively, the recent Contrastive Language-Image Pretraining (CLIP) model \cite{clip} has shown promise in zero-shot recognition across various vision tasks, such as detecting unknown objects \cite{f-vlm} and identifying out-of-distribution images \cite{clip-ood}. Building on this paradigm, WinCLIP \cite{winclip} used text prompts for anomaly measurement, significantly improving over other category-agnostic methods in a zero-shot AD setup and extending the capabilities of the CLIP model \cite{clip}. Subsequent approaches like \cite{anoclip, aprilgan, randomwordclip, anomalyclip} further enhanced ZSAD capabilities. Recent works \cite{aprilgan, anomalyclip} have begun fine-tuning the pretrained CLIP model with auxiliary anomalous image and, testing it on the target datasets. Alternatively, AnoCLIP \cite{anoclip} introduced a test-time adaptation (TTA) module to alter the visual representation space of the CLIP model. These studies underscore the importance of adding training parameters to the pretrained CLIP model to strengthen its anomaly localization ability. However, solely incorporating semantic information from text prompts may not fully exploit the potential of large vision-language models. Since the vision-language space isn't perfectly aligned, many visual anomalies implicitly defined in the visual distribution remain uncovered. Additional visual references need to be incorporated to assist the language-based ZSAD, especially for misplaced objects, rare-seen objects, and complicated scenes, whose anomaly information is usually hard to obtain from large pretrain datasets.


To address these issues, we propose a novel framework (see \cref{fig: methodology}) that utilizes a pair of unlabelled images during testing. Our framework comprises a pretrained CLIP model, a test-time adaptation module, and an input path for image pairs to leverage the additional visual reference information into the language-vision AD. The anomaly score of a query image depends not only on its textual zero-shot score but also on the score derived from its randomly paired reference image.
Additionally, we enhanced the model's AD capability by adding a TTA module involving pseudo anomaly synthesis to improve the agnostic ability to locate anomalies.

In summary, our contributions are threefold:
\begin{itemize}
    \item We propose a novel ZSAD method that processes a pair of images, enhancing existing CLIP-based AD methods. This approach incorporates an additional reference image, operates without the need for further training and significantly boosts AD performance.
    \item We developed a TTA module that includes pseudo anomaly synthesis methods adopted from DRAEM \cite{draem}, effectively refining the AD capabilities of the pretrained CLIP model.
    \item Comprehensive experiments on MVTecAD \cite{mvtec} and VisA \cite{visa} reveal that our methods achieve comparable performance with current SOTA ZSAD methods in both anomaly classification and anomaly localization.
\end{itemize}

\section{Related Work}
\label{sec: relate}

\subsection{Vision-Language Models}
Advancements in data scale have led to significant strides in pretrained visual language models \cite{virtex, uniter, clip, flamingo}, which demonstrate remarkable proficiency in a variety of downstream tasks \cite{coop, cocoop, maple, self_regulate_maple, medicalclipdetection, clip-ood, clipasso}. A prime example is the CLIP model, through contrastive vision-language pre-training on a diverse array of internet-sourced image-text pairs, it exhibits exceptional generality and adaptability. This model is particularly adept at zero-shot inference, displaying a superior capacity for recognizing images beyond its training data. Recent explorations have extended the zero-shot capabilities of CLIP models to tasks like open-vocabulary semantic segmentation, achieved by harnessing intrinsic dense features \cite{denseclip, maskclip, clipsurgery}. Additionally, efforts to optimize CLIP's recognition performance have been fruitful, focusing on areas such as prompt engineering \cite{coop, cocoop}, adapter modules \cite{tipadapter, clipadapter}, and additional training for enhanced vision-language alignment \cite{maple, self_regulate_maple}. Importantly, CLIP's inherent ability to detect out-of-distribution data without additional training has catalyzed its application in zero-shot anomaly classification and localization.

\subsection{Anomaly Detection and Localization}
The essence of most existing AD methods lies in modeling the distribution of normal samples to detect anomalies \cite{fanogan, mkd, rd, draem}. Particularly, reconstruction-based approaches, such as those using autoencoders \cite{ganomaly, skipganomaly, memae} and GANs \cite{fanogan}, have been widely adopted. These methods leverage the reconstruction capabilities of anomaly-free images, hypothesizing that abnormalities will manifest as discrepancies in the reconstructed output. Additionally, discriminative models have been developed, often with the aid of synthetic anomalies \cite{draem, cutpaste}. Moreover, large pretrained models like ImageNet \cite{imagenet} have been found to be robust in anomaly detection, by comparing features embedded from normal images \cite{padim, patchcore, spade}. To bridge the gap between natural image pretraining and industrial application, further adaptations have been employed, such as student-teacher knowledge distillation \cite{mkd, rd} and normalizing flows \cite{glow, cflowad}.

A significant challenge in this field is the development of unified models capable of detecting anomalies across multiple classes \cite{uniad, testtime_prompt_uniad, structural_hanqiu}. In the context of zero-shot settings, MuSc \cite{musc} innovatively utilizes unlabelled images from the test set as references. Alternatively, the WinCLIP model \cite{winclip} utilizes CLIP \cite{clip} for prompt-guided AD. Building on this, AnoCLIP \cite{anoclip} enhances localization representation and implements V-V attention introduced in \cite{clipsurgery}. However, vision-language models such as CLIP are primarily trained to align with the class semantics of foreground objects rather than the abnormality/normality in the images, and as a result, their generalization in understanding the visual anomalies is restricted, leading to weak ZSAD performance. Existing zero-shot prompt-guided AD models often lack robust visual representation as a basis for detecting anomalies. Addressing this, methods such as \cite{aprilgan, anomalyclip} propose fine-tuning the pretrained CLIP model with auxiliary images for cross-set training/validation. In light of the limitations inherent in current CLIP-based anomaly detection models and the essential role of visual references, we propose an innovative, training-free method that inference on pairs of images. In this approach each image acts as a visual reference for the other, significantly enhancing the accuracy and effectiveness of anomaly detection.

\section{Methodology}
\label{sec: methodology}
\begin{figure}[tb]
    \centering
    \includegraphics[width=0.95\textwidth]{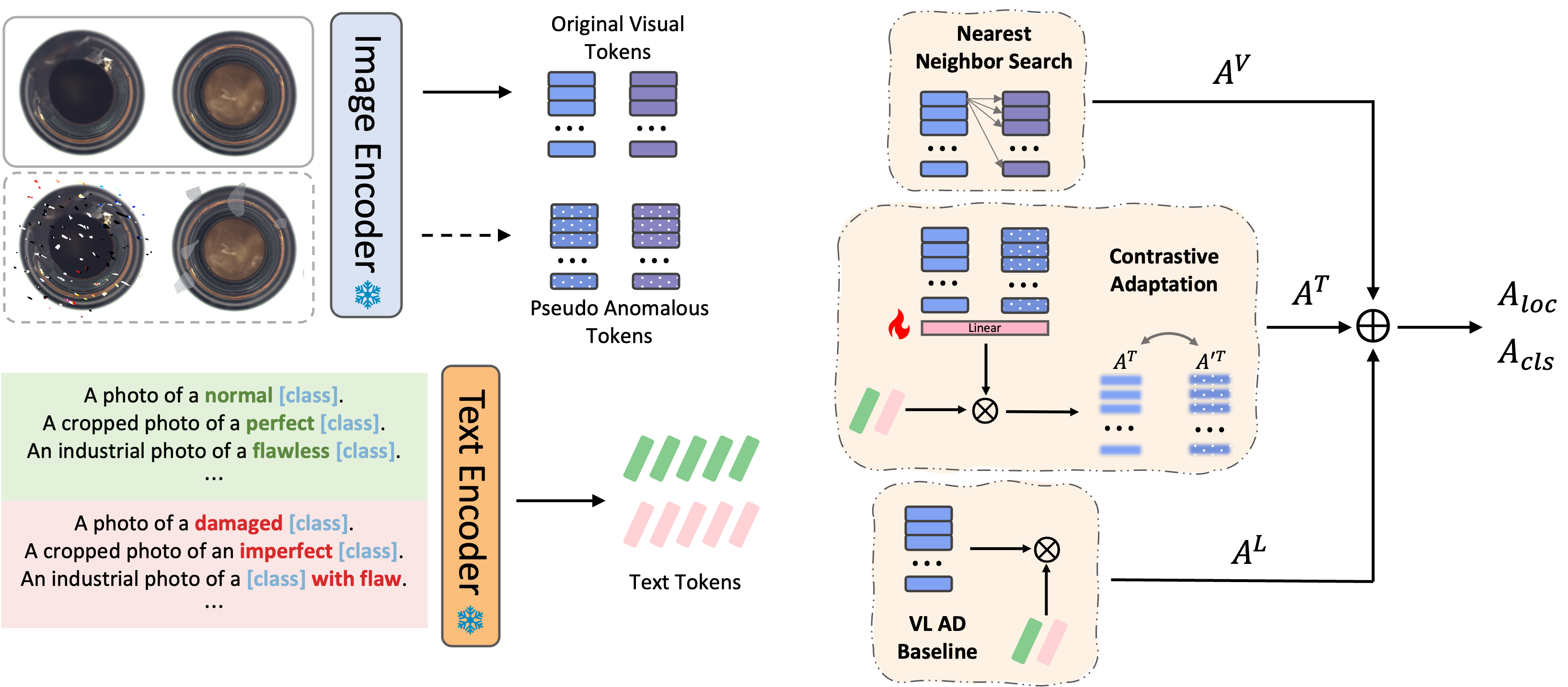}
    \caption{Overview of our framework for Dual Image Enhanced CLIP. The left part shows the feature extraction process from the vision and text encoder, and the right section shows the inference process. The snowflake denotes the modules are frozen, and the flame icon represents trainable modules.}
    \label{fig: methodology}
\end{figure}

In this section, we first introduce the CLIP-based baseline model for zero-shot anomaly classification and localization. Following this, we delve into details of our dual-image enhancement model. Lastly, we specify our test-time adaptation mechanism to refine the model's AD capability. Fig. \ref{fig: methodology} provides a comprehensive overview of our framework.

\subsection{CLIP for Zero-shot Anomaly Detection}
\label{subsec: clip ZSAD}

CLIP's zero-shot visual recognition, trained on a multi-million image-text pair dataset, aligns images with textual descriptions through a visual encoder and a text encoder. These encoders respectively transform images and text prompts (e.g., \textit{a photo of a [class]}) into visual and text tokens in a shared feature space. The model's ability to compare these tokens via cosine similarity allows it to identify class concepts within images.

For anomaly detection, CLIP utilizes semantic concepts of  ``normal'' and ``anomalous'' states. Multiple prompts with varied descriptors (like ``perfect'', ``broken'', \etc.) are used to create averaged text tokens representing these states, $t_n$ and $t_a$ for normal and anomalous text tokens, respectively. Anomaly score for an image is computed based on the similarity between its visual token and these averaged text tokens. Specifically, given a text prompt and the corresponding class token $v$, the sample-level anomaly score $A^L_{cls}$ is computed as:

\begin{equation}
A^L_{cls} = F(v, t_a, t_n) = \frac{\exp(\langle v, t_a \rangle)/\tau}{\exp(\langle v, t_n \rangle/\tau)) + \exp(\langle v, t_a \rangle/\tau)}
  \label{eq:important}
\end{equation}
where $\tau$ is the temperature hyperparameter. Note that no visual information is injected into the model, but rather unknown anomalies are detected through the powerful open-world generalization of CLIP.

The computation is extended from global visual embeddings to patch-level visual embeddings to derive the corresponding segmentation maps $A^L_{loc} \in \mathbb{R}^{H\times W}$, the final layer of the visual encoder has a set of patch tokens $p_{(j, k)} \in \mathbb{R}$ that potentially contain image local information in the patch level. For a patch token $p_{(j,k)}$, the local anomaly score is computed as:


\begin{align}
    A^L_{loc} 
    &= \left\{ F(p_{(j,k)}, t_a, t_n) \right\}_{j=0,k=0}^{h-1,w-1} \\
    & = \left\{ \frac{\exp(\langle p_{(j,k)}, t_a \rangle / \tau)}{\exp(\langle p_{(j,k)}, t_n \rangle / \tau) + \exp(\langle p_{(j,k)}, t_a \rangle / \tau)} \right\}_{j=0,k=0}^{h-1,w-1}
  \label{eq:important}
\end{align}

However, since CLIP was primarily trained to align the class tokens with the text token for global classification, there's a lack of alignment between local patch tokens and text embeddings that leads to limited performance in segmenting anomalous regions. Hence, after iterative explorations \cite{clipsurgery, anoclip, anomalyclip}, V-V attention was adopted to produce the local-aware patch tokens.


In original Q-K-V attention, the attention score can be disproportionately influenced by specific tokens, leading to a representation that is disturbed by unrelated local features, which can weaken the model's localization ability to detect anomalies. The V-V attention mechanism is proposed as an alternative that enhances the local features without additional training. This novel attention mechanism replaces the queries and keys with values. 

\begin{equation}
    V^l = Proj.(Attention(V^{l-1}, V^{l-1}, V^{l-1})) + V^{l-1}
\end{equation}

By focusing on self-attention within the values themselves, V-V attention avoids bias introduced by the query and key interactions in Q-K-V attention. It reduces the disturbance caused by other tokens, ensuring that each value contributes significantly to its own representation. As a result, attention maps produced by V-V attention exhibit a pronounced diagonal pattern, indicating that each token predominantly attends to itself, thereby preserving its local information.

In our model, the architecture remains the same with AnoCLIP \cite{anoclip}, which follows a 2-way forward path. The original Q-K-V attention path was kept to produce the class token, which was used to calculate sample level anomaly score $A^L_{cls}$. The patch tokens used for localization score $A^L_{loc}$ are all computed by the V-V attention path.

\subsection{Dual Image Feature Enhancement}
\label{subsec: dual image}

\begin{figure}
    \centering
    \includegraphics[width=\textwidth]{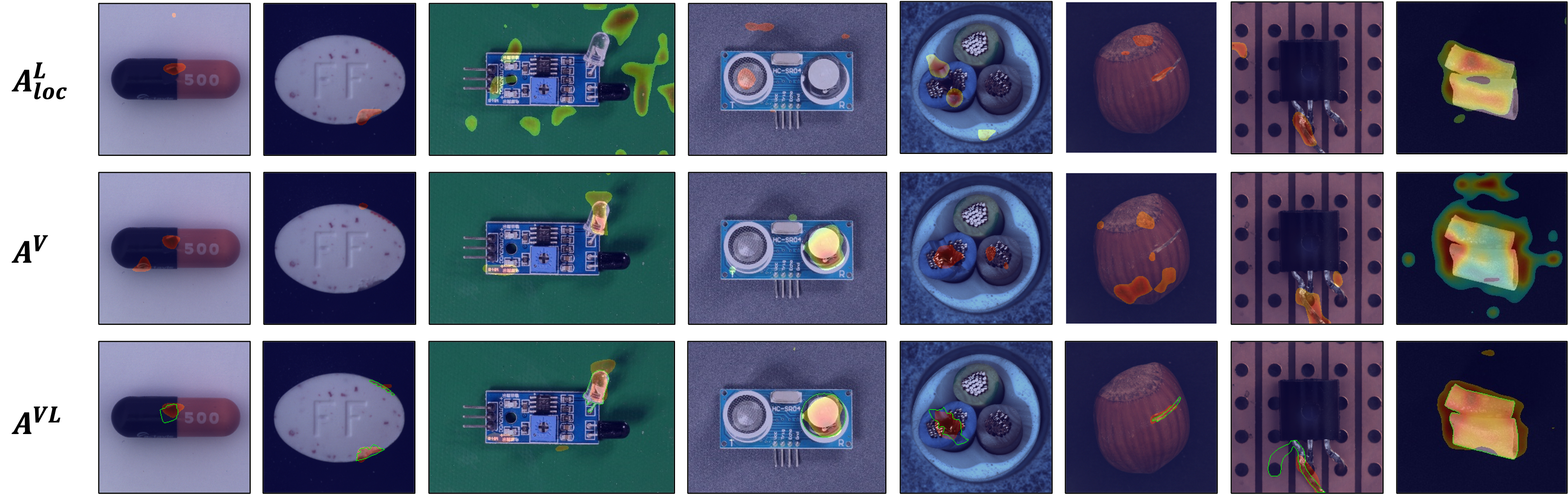}
    \caption{Qualitative illustration of the comparison with AD results on MVTecAD and VisA. The top row illustrates the result solely using textual information. The middle row depicts detection results through paired queries' visual feature comparison. The bottom row showcases more robust results achieved by integrating both language and visual features, and the ground truth is marked with green boundaries.}
    \label{fig: single vs pair}
\end{figure}

As shown in \cref{fig: methodology}, we proposed a novel approach that inputs a pair of images in test-time. Unlike previous CLIP-based AD works \cite{winclip, anoclip, anomalyclip, randomwordclip} which predominantly rely on text prompts for inference, we incorporate additional visual information to facilitate a more comprehensive joint vision-language prediction. To highlight the effectiveness of our approach, we provide a comparative analysis in \cref{fig: single vs pair}.

\cref{fig: single vs pair} demonstrates a significant observation: the exclusive dependence on either textual or visual information alone proves inadequate for the accurate detection of certain anomalies. The limitation in leveraging text stems from the constraints inherent in utilizing state descriptions within prompts, with terms like ``broken'' or ``damaged'' falling short of encapsulating the full spectrum of potential anomalies. Making inferences on a single image invites biases and misinterpretations, emphasizing a more extensive visual intra-class diversity to form a baseline for normalcy. For instance, consider the ``PCB'' example in column 3 of \cref{fig: single vs pair}, where a misplaced LED is an anomaly; however, using text description alone is insufficient for its detection. Moreover, logical anomalies, global distortions, rare objects, or complicated scenes are more challenging to discern from by text-based method. This emphasizes the importance of incorporating additional visual context for a varied example to enhance the detection accuracy. Conversely, relying solely on pairwise visual comparisons also presents limitations, as the reference image could itself be anomalous. Consequently, this paves the way for an integrated approach that combines textual and visual data to overcome these challenges.  As depicted in \cref{fig: single vs pair}, employing dual-image inputs within the CLIP-based framework mitigates these issues, contributing to improved anomaly localization accuracy.

In response to these findings, our framework introduces a novel strategy that capitalizes on both textual and visual features. This is achieved by a unique process of randomly selecting pairs of test images to serve as the query and reference images. For these image pairs, we extract patch tokens, denoted as \( q_{(j,k)} \) for the query and \( r_{(m,n)} \) for the reference. These patch tokens form the basis for the pairwise visual feature comparison.

In this pairwise feature comparison strategy, each patch token \( q_{(j,k)} \) from the query image undergoes a nearest neighbour search with the patch tokens from the reference image, effectively using the latter as a memory repository. The anomaly score for each query patch token \( q_{(j,k)} \) is determined by calculating its cosine similarity with all reference patch tokens set \( \mathcal{S}_r = \{ r_{(m,n)} \mid m \in \{1, \ldots, H\}, n \in \{1, \ldots, W\} \} \). The maximum similarity score, indicating the minimum deviation, is then designated as the anomaly score for the query patch:


\begin{equation}
    A^{V}_{(j,k)} = \min_{r_{(m,n)} \in \mathcal{S}_r} \left( 1 - \text{sim} \left( q_{(j,k)}, r_{(m,n)} \right) \right)
\end{equation}

In the equation above, \( A^{V}_{(j,k)} \) represents the visual reference anomaly score of the query patch \( q_{(j,k)} \),  the overall anomaly score $A^V \in \mathbb{R}^{H\times W} $ for the query image, is a composition of all the patch scores across the entire image. $\text{sim}$ represents the cosine similarity between the patch tokens of the two samples.

As a result, the vision-language joint anomaly score can be computed as:
\begin{equation}
    A^{VL}_{loc} = A^{V} + A^{L}_{loc}
\end{equation}

\subsection{Test-Time Adaption with Pseudo Anomaly Synthesis}
\label{subsec: tta}

\begin{figure}[t]
    \centering
    \includegraphics[width=\textwidth]{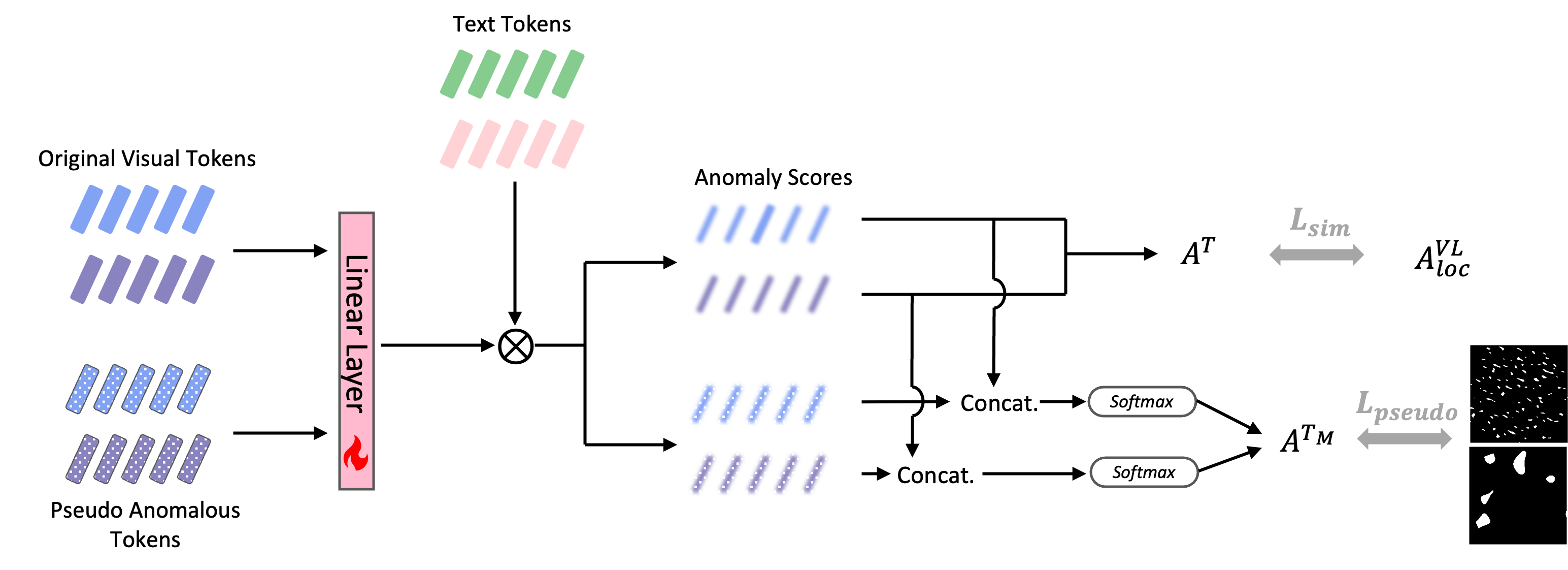}
    \caption{Workflow of the test-time adaptation module. The module inputs patch tokens through a linear layer, aligning predictions on the adapted token with the zero-shot vision-language joint anomaly score. Pseudo-anomalous samples are compared with original samples to predict pseudo-anomaly masks. The flame icon denotes trainable components. $A^{T_M}$ denotes the prediction for the pseudo anomalies.}
    \label{fig: tta}
\end{figure}

As the visual-language alignment needed to be refined for AD, we proposed a test-time adaptation module to boost the CLIP-based model's AD capabilities. Our TTA module is achieved through a linear adapter, as depicted in \cref{fig: tta}. For the original image, we utilize the pseudo anomaly synthesis technique from DRAEM \cite{draem} to introduce image corruptions. DRAEM creates random-shaped pseudo anomaly masks using Perlin noise \cite{perlin} and overlays textures from \cite{dtd} onto the original image at masked locations. The resultant pseudo-anomalous patch tokens, denoted as \( q'_{(j,k)} \in \mathbb{R} \), encapsulate pseudo-anomalous features.

The online adaptation of the original and synthesized patch tokens is mathematically represented as:
\begin{align}
     q^T_{(j,k)} &= \frac{1}{2} \left( G(q_{(j,k)}) + q_{(j,k)} \right) \\
     q'^T_{(j,k)} &= \frac{1}{2} \left( G(q'_{(j,k)}) + q'_{(j,k)} \right)
\end{align}
Here $G(\cdot)$ denotes the linear computation. Subsequently, these adapted patch tokens are aligned with text tokens to compute the anomaly score:
\begin{equation}
    A^T = \left\{ F(q^T_{(j,k)}, t_a, t_n) \right\}_{j=0,k=0}^{h-1,w-1}
    \label{eq: tta anomaly score}
\end{equation}

To optimize the weights of the linear layer, we establish self-supervised tasks using pseudo anomalies. For the original and adapted patch tokens $q^T_{(j,k)}$, $q'^T_{(j,k)}$ from queries, we design two discriminative self-supervised tasks for TTA:

(1) For predicting pseudo anomaly masks \( M_a \), we define the \( L_{\text{pseudo}} \) loss as:
\begin{equation}
L_{\text{pseudo}} = \frac{1}{|\mathcal{S}_a|} \sum_{(j,k) \in \mathcal{S}_a} \left( -M_a \cdot \log\left(\frac{\exp(A'^T)}{\exp(A^T) + \exp(A'^T)}\right) \right)_{j,k}
\end{equation}
Here, \( \mathcal{S}_a \) represents the set of indices \( (j,k) \) where \( M_a \neq 0 \), indicating regions augmented by the pseudo mask \( M_a \). \( L_{\text{pseudo}} \) prompts the adapter to retain abnormal features and recognize pseudo anomalies, aiding in the subtle detection of real anomalies.

(2) To encourage the adapter to preserve normal features and uphold general anomaly detection capabilities, we utilize the similarity loss \( L_{\text{sim}} \) to ensure that adapted anomaly scores \( A^T \) are consistent with the zero-shot vision-language joint localization:
\begin{equation}
    L_{\text{sim}} = \text{sim} \left(A^{VL}_{loc}, A^T\right)
\end{equation}

The aggregate learning objective to train our adapter is \( L = L_{\text{pseudo}} + \beta L_{\text{sim}} \). This TTA process is efficient and does not require any training data or annotation. Finally, the overall anomaly classification and localization score for the query image should be computed as:
\begin{align}
    A_{loc} &= \lambda_1 A^{V} + \lambda_2 A^T \\
    A_{cls} &=  \lambda_3 A^{L}_{det} + \lambda_4 \max_{j,k} A^{V} + \lambda_5 \max_{j,k} A^{T}
\end{align}

\section{Experiment}
\label{sec: experiment}

\subsection{Experimental Setup} 
    \subsubsection{Datasets.}

In our study, we conducted experiments using the MVTecAD \cite{mvtec} and VisA \cite{visa} datasets. Both of these datasets offer a wide array of subsets featuring various objects and textures. MVTecAD includes high-resolution images with dimensions varying from $700^2$ to $1024^2$, while the VisA comprises rectangular images with resolutions around $1.5K\times1K$, each accompanied by corresponding anomaly ground truth masks. Specifically, MVTecAD encompasses 5 texture categories and 10 object categories, whereas VisA is composed of 12 subsets, each dedicated to different objects. In this paper, we exclusively utilized the test dataset to evaluate zero-shot anomaly classification and localization, without the acquisition of additional datasets.


\subsubsection{Metrics.} \label{subsec: metrics}
    We assess the efficacy of our model by utilizing the Area Under Receiver Operator Characteristics (AUROC) image-level AUROC is used for anomaly detection, and pixel-level AUROC is measured for evaluating anomaly localization. However, the metric is dominated by a large number of normal pixels and is thus kept high despite false detections. We thus additionally report the F1Max score and Area Under Precision-Recall (AUPR) as a balanced calculation of the precision and recall to overcome the class imbalance. In addition to that, we compute the Per-Region-Overlap (PRO) to measure anomaly localization, which weights each connected component within the ground truth of varying sizes equally, making it more robust than simple pixel measurement.
    
\subsubsection{Implementation.} \label{subsec: implementation}
We adopt ViT-B-16+ \cite{dosovitskiy2021image} as the visual encoder and the transformer \cite{vaswani2023attention} as the text encoder by default from the public pretrained CLIP model \cite{openclip}. For the text encoder, following previous work of AnoCLIP \cite{anoclip}, employing $22$ base templates collected from CLIP \cite{clip}, $7$ pairs of state prompts, and $4$ domain-aware prompts to generate sufficient prompts. All prompts are listed in the supplement. We adhered to the data preprocessing pipeline outlined in OpenCLIP \cite{openclip} for both MVTecAD and VisA benchmarks, standardizing image sizes to $(240, 240)$. Regarding the scoring coefficients, we configured $\lambda_1, \lambda_3, \lambda_4, \lambda_5$ to $1$, and set $\lambda_2$ to $1.5$. For TTA, we use the AdamW \cite{adam} optimizer and set the learning rate to $0.001$, $\beta=0.5$, the adaptor is optimized with $2$ training steps. We report the mean and variance of the results over $6$ random seeds.

\subsection{Performance}
\label{subsec: performance}

\cref{tab: combined_anomaly_detection_comparison} presents the performance of zero-shot anomaly detection on MVTecAD and VisA datasets. Our proposed method is compared with prior ZSAD based works, including CLIP \cite{clip}, WinCLIP \cite{winclip}, AnoCLIP \cite{anoclip}, and MuSc \cite{musc}. Notely, MuSc utilized the entire test set for visual reference, aligning with transductive rather than inductive learning. For a fair comparison, we adapted MuSc to our pairwise image setting and used ViT-B-16+ as the backbone, denoted as MuSc-2. From the table, our proposed methods exhibit exceptional performance, significantly outperforming AnoCLIP by margins of $2.2\%, 6.0\%, 6.2\%$ in AUROC, F1Max, and PRO for anomaly localization. We also achieve advancements in anomaly classification, surpassing other methods by substantial margins. This trend of exceptional performance is consistent on the VisA dataset. Qualitative results for ZSAD are further detailed in \cref{fig: single vs pair}, illustrating our model's capacity to effectively classify and localize the anomalies across varied samples.

Additionally, \cref{tab: comparison with SOTA} presents a comparison of our method against other AD models by AUROC scores on MVTecAD. Here, we categorize current zero-shot anomaly detection methodologies into three paradigms: Auxiliary Data approaches that utilize additional anomalous data for training, Transductive Learning Methods that infer from the extensive portion of the test set, and Inductive Learning Approaches that assess each query independently, without prior knowledge of the overall distribution. Our method, an exemplar of the inductive approach, outperforms Auxiliary Data Approaches without requiring exposure to anomalies during training.

Contrasting our method with MuSc, the state-of-the-art transductive learning setting of ZSAD, MuSc requires accessing the entire test set distribution before inference, making it highly dependent on the data distribution, and potentially limiting in real-world applications like online real-time inference. In \cref{tab: combined_anomaly_detection_comparison}, our method surpasses MuSc's performance on the pair-image setting, especially in anomaly classification. This derives advantages from the utilization of the class token in CLIP embeddings, and emphasizes the efficiency and robustness of our joint language-vision prediction.

\begin{table}[]
\centering
\caption{Zero-shot Anomaly Localization (AL) and Anomaly Classification (AC) on MVTecAD and VisA datasets. Bold indicates the best performance and underline indicates the runner-up unless other noted. MuSc-2 denotes inference  with 2 images.}
\label{tab: combined_anomaly_detection_comparison}
\resizebox{\textwidth}{!}{%
\begin{tabular}{@{}ccccccccccccc@{}}
\toprule
\multirow{3}{*}{Methods}& \multicolumn{6}{c}{MvTecAD} & \multicolumn{6}{c}{VisA} \\
\cmidrule(lr){2-7} \cmidrule(lr){8-13}
 & \multicolumn{3}{c}{AL} & \multicolumn{3}{c}{AC} & \multicolumn{3}{c}{AL} & \multicolumn{3}{c}{AC} \\ 
\cmidrule(lr){2-4} \cmidrule(lr){5-7} \cmidrule(lr){8-10} \cmidrule(lr){11-13}
& AUROC & F1Max & PRO & AUROC & F1Max & AP & AUROC & F1Max & PRO & AUROC & F1Max & AP \\ 
\midrule
CLIP \cite{clip} & 19.5 & 6.2 & 1.6 & 74.0 & 88.5 & 89.1 & 22.3 & 1.4 & 0.8 & 59.3 & 74.4 & 67.0 \\
WinCLIP \cite{winclip} & 85.1 & 31.7 & 64.6 & 91.8 & 91.9 & 96.5 & 79.6 & 14.8 & 59.8 & 78.1 & 79.0 & 81.2 \\
AnoCLIP \cite{anoclip} & 90.6 & 36.5 & 77.8 & 92.5 & 93.2 & \textbf{96.7} & 91.4 & 17.4 & 75.0 & 79.2 & 79.7 & 81.7 \\
MuSc-2 \cite{musc} & 92.4 & 41.2 & 76.5 & 81.7 & 89.1 & 90.3 & 92.6 & \textbf{26.7} & 63.2 & 69.4 & 75.1 & 73.3 \\
\textbf{Ours} & {\ul 92.6} & {\ul 41.8} & {\ul 82.6} & {\ul 93.1} & {\ul 94.0} & {\ul 96.6} & \textbf{94.8} & 23.5 & {\ul 78.6} & {\ul 82.6} & \textbf{81.0} & {\ul 84.2} \\
\textbf{Ours+} & \textbf{92.8} & \textbf{42.5} & \textbf{84.0} & \textbf{93.2} & \textbf{94.1} & \textbf{96.7} & {\ul 94.2} & {\ul 24.1} & \textbf{79.7} & \textbf{82.9} & {\ul 80.9} & \textbf{84.7} \\ 
\bottomrule
\end{tabular}%
}
\end{table}

\begin{table}[]
\caption{Comparative analysis of AD methods in Full-shot and Zero-shot settings.}
\label{tab: comparison with SOTA}
\centering
\resizebox{\textwidth}{!}{%
\begin{tabular}{@{}ccccccccc@{}}
\toprule
\multirow{2}{*}{Setting} & \multicolumn{2}{c}{\multirow{2}{*}{Full Shot}} & \multicolumn{6}{c}{Zero-shot} \\ \cmidrule(lr){4-9}
& & & \multicolumn{2}{c}{Aux. Data} & \multicolumn{1}{c}{Transd.} & \multicolumn{3}{c}{Induct.} \\
\cmidrule(lr){1-1} \cmidrule(lr){2-3} \cmidrule(lr){4-5} \cmidrule(lr){6-6} \cmidrule(lr){7-9}
Methods & DRAEM \cite{draem} & UniAD\cite{uniad} & AprilGAN \cite{aprilgan} & AnomalyCLIP\cite{anomalyclip} & MuSc \cite{musc} & WinCLIP \cite{winclip} & AnoCLIP \cite{anoclip} & Ours \\
\midrule
A.C. & 88.1 & 96.5 & 87.6 & 91.5 & 97.8 & 91.2 & 92.5 & 93.2$\pm$0.8 \\
A.L. & 87.2 & 96.8 & 86.1 & 91.1 & 97.3 & 85.1 & 90.6 & 92.8$\pm$0.2 \\
\bottomrule
\end{tabular}%
}
\end{table}

\section{Ablation Studies}
\label{sec: ablation}

\subsection{Reference Images Quantity}
\label{subsec: reference images}

\begin{figure}[tb]
  \centering
  \begin{subfigure}{0.45\textwidth}
    \includegraphics[width=\textwidth]{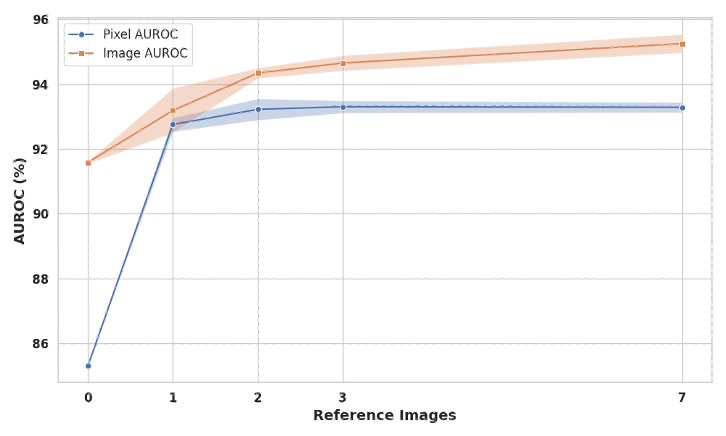}
    \caption{AUROC on MVTecAD with the increasing number of reference images.}
    \label{fig: batchsize}
  \end{subfigure}
  \hfill
  \begin{subfigure}{0.50\textwidth}
    \includegraphics[width=\textwidth]{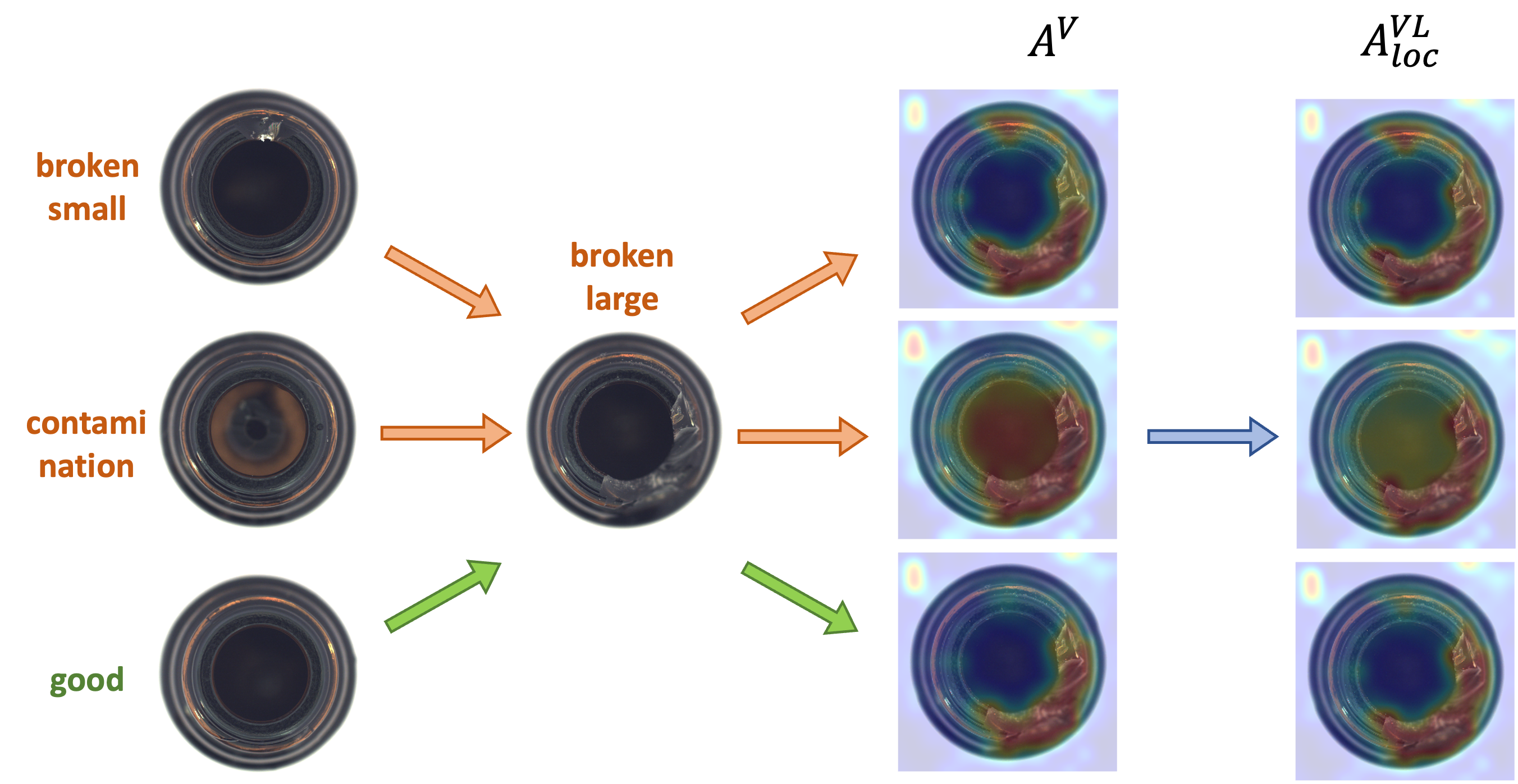}
    \caption{Impact of Reference Image Selection, illustrating variance in anomaly score on the choice of normal samples and various anomalous samples.}
    \label{fig: reference choice}
  \end{subfigure}
  \caption{Ablation studies on the influence of reference images.}
  \label{fig: reference images}
\end{figure}

From \cref{fig: batchsize}, a clear trend is observed: with the increasing number of reference images, the anomaly detection performance improves, reflected from both pixel-level and sample-level AUROC. This supports the hypothesis that unlabelled images can still offer valuable comparative references for anomaly identification.

Significantly, the most pronounced performance leap occurs when the number of reference images is increased from $0$ to $1$, indicating that even a single reference image can substantially improve the model's AD ability. However, the subsequent performance gains from $1$ to $7$ reference images are present but exhibit diminishing returns. 
This trend implies that considering the factors of inference time and memory efficiency, utilizing a large pool of reference images might not always be feasible or optimal, particularly in real-world applications where access to a wide array of suitable samples is often limited. In this context, a pairwise approach emerges as a balanced solution, optimizing the trade-off between improved detection performance and computational resource efficiency.
Moreover, the observed differences in the degree of improvement between pixel AUROC and sample AUROC point that while visual details are paramount in pinpointing anomalies, they may be less influential in the broader context of classifying an entire sample as anomalous. 

\subsection{The Choice of Pairing Samples}

Since the reference images are randomly sampled from unlabelled images, they can either be normal or anomalous. This leads to the question: how does an anomalous reference image affect the precision of anomaly detection?

\cref{fig: reference choice} illustrates how different reference images can significantly influence the anomaly score. In the case of the ``bottle'', it is evident that using a normal reference image generally guarantees the accuracy of anomaly detection, as it provides a clear baseline for identifying outliers. Conversely, when the reference image contains anomalies, such as breaks or contamination, these imperfections can misleadingly provide a false reference for the query image, erroneously highlighting the reference image's damaged region in the query. This phenomenon suggests that the abnormal condition of the reference image can ``pollute'' the anomaly score of the query image.

Interestingly, integrating textual cues with visual data can mitigate this negative effect. By leveraging textual features from prompts, the model can effectively counter the false prediction associated with the anomalous references. As depicted in \cref{fig: reference choice}, the joint predictions that combine both visual and language information exhibit a notable increase in accuracy, underscoring the potential of language-vision joint anomaly detection.


\subsection{Test-Time Adaption Module}
\label{subsec: tta}

\begin{figure}[tb]
  \centering
  \begin{subfigure}[b]{0.59\textwidth}
    \centering
    \includegraphics[width=\textwidth]{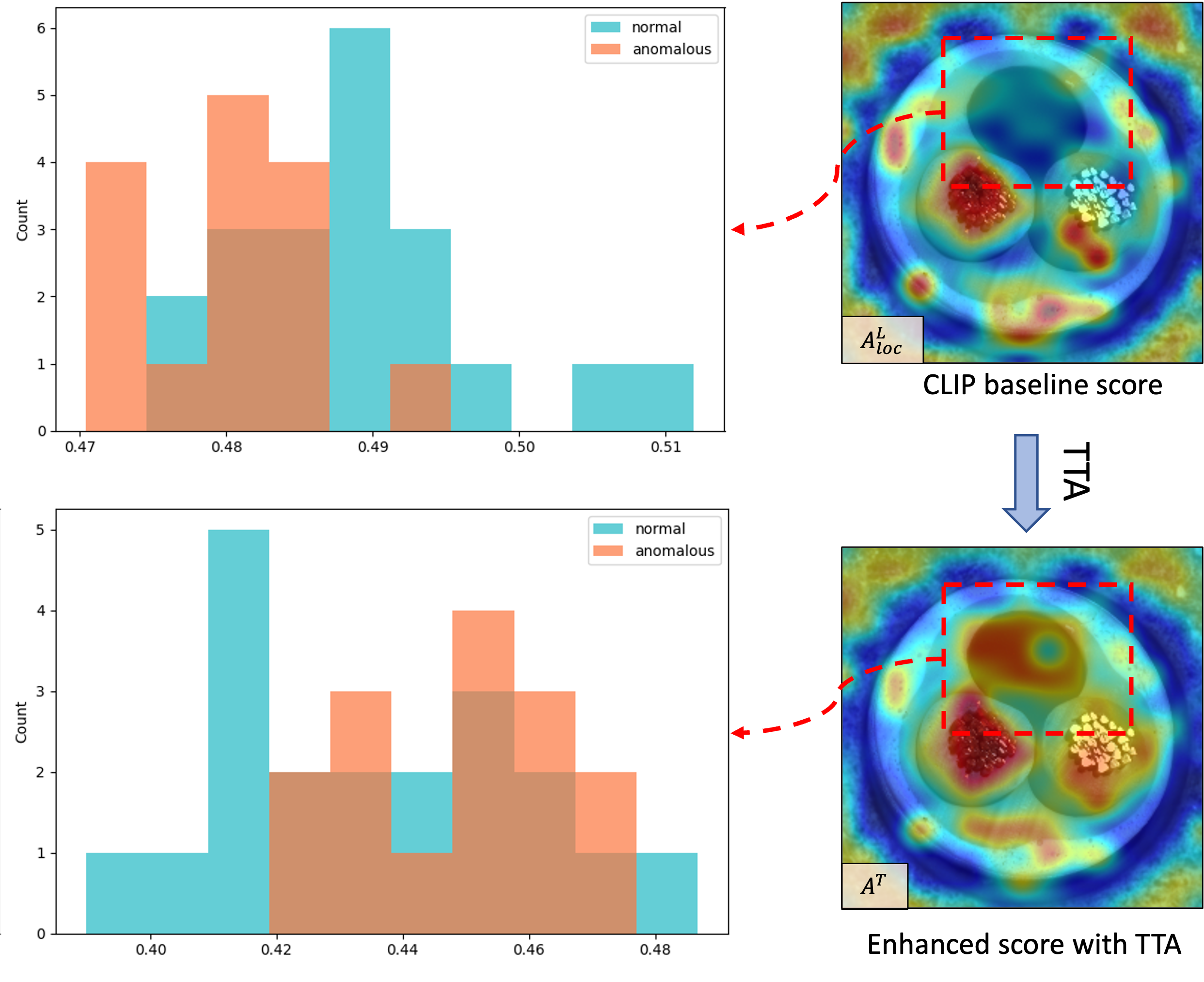}
    \caption{Left: Histogram of the TTA anomaly score from the highlighted red box region. Right: Heatmap of the anomaly score for ``missing cable'' before and after adaptation.}
    \label{fig: histogram tta}
  \end{subfigure}
  \hfill
  \begin{subfigure}[b]{0.37\textwidth}
    \centering
    \begin{subfigure}{\textwidth}
      \centering
      \includegraphics[width=\textwidth]{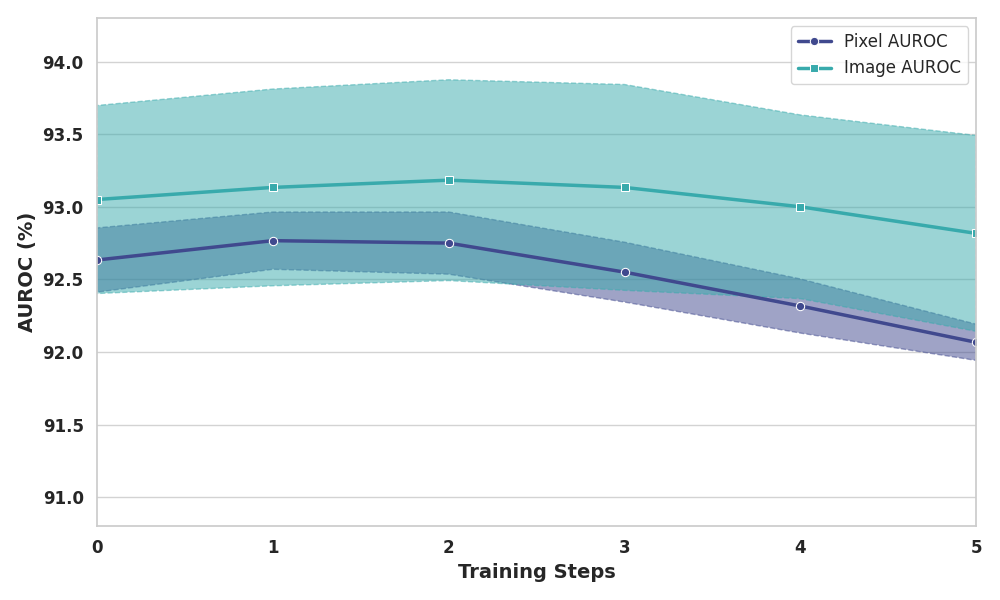}
      \caption{AUROC on MVTecAD with different TTA training steps.}
      \label{fig: training steps AUC}
    \end{subfigure}
    \\
    \begin{subfigure}{\textwidth}
      \centering
      \includegraphics[width=\textwidth]{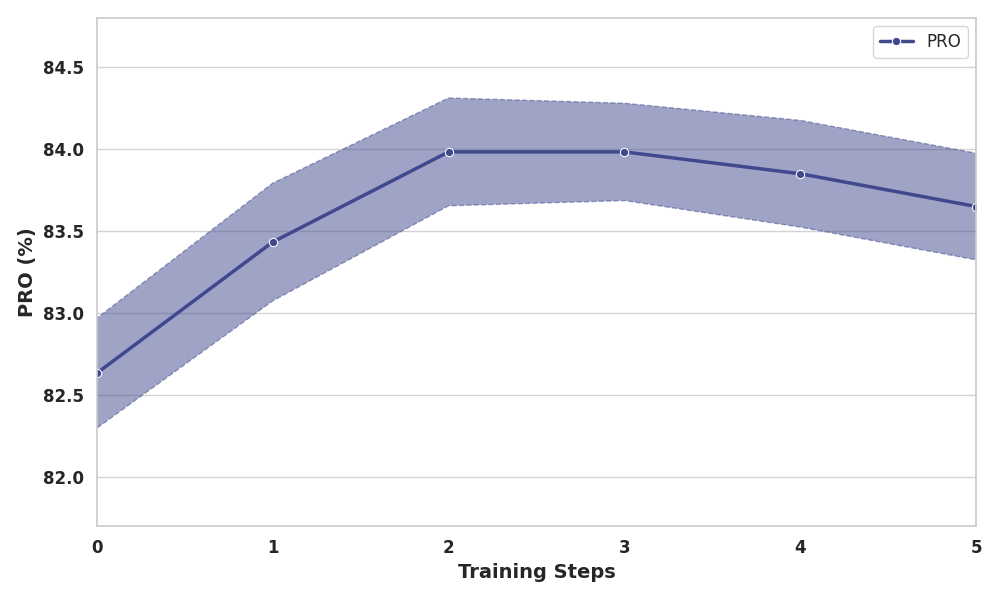}
      \caption{PRO on MVTecAD with different TTA training steps.}
      \label{fig: training steps PRO}
    \end{subfigure}
    \label{fig: tta ablations}
  \end{subfigure}

  \caption{Ablation studies on test-time adaptation.}
  \label{fig:short}
\end{figure}

We also studied the impact of various training steps  on model performance, \cref{fig: tta ablations} demonstrates the AUROC and PRO for training steps ranging from 1 to 6. We can see both the pixel and sample AUROC and PRO scores reach the optimal when the training step is set to 2, and start to decrease. Therefore, we opted for a $\text{training step} = 2$. In \cref{tab: ablation on TTA}, we showcase the performance enhancements by the TTA module. Here we take the text-only approach as the baseline. The table shows that implementing the TTA module on the baseline yields a notable increase in performance. When the TTA module operates alongside a paired reference image, the results are further amplified. In this scenario, the AUROC for AL climbs to $92.8\%$, and AC reaches $93.2\%$. Further, the improvement in the F1Max and PRO indicates a more balanced and effective model, particularly in terms of its localization capabilities. The influence of the TTA module is further presented in \cref{fig: histogram tta}, where a marked distinction in anomaly scores between normal and anomalous patches is observed post-adaptation. Prior to adaptation, the ``missing cable'' region was not adequately identified, and the adaptation process leads to a refined alignment between visual perception and language context, resulting in superior AL and AC performance.

\begin{table}[]
\caption{Ablation studies on the TTA module.}
\label{tab: ablation on TTA}
\centering
\begin{tabular}{@{}cccc@{\hspace{6pt}}c@{\hspace{6pt}}c@{\hspace{6pt}}c@{\hspace{6pt}}c@{\hspace{6pt}}c@{}}
\toprule
& & & \multicolumn{3}{c}{A.L.} & \multicolumn{3}{c}{A.C} \\
\cmidrule(lr){4-6} \cmidrule(lr){7-9}
\multirow{2}{*}{} & \multirow{-2}{*}{$+A^V$} & \multirow{-2}{*}{$+\text{TTA}$} & AUROC & F1Max & PRO & AUROC & F1Max & AP \\
\midrule
\multirow{5}{*}{base.} & & & $85.3\pm0.0$ &  $29.1\pm0.1$ & $71.8\pm0.4$& $91.6\pm0.0$ & $92.9\pm0.1$ & $96.4\pm0.0$ \\
& & \checkmark & $88.7\pm0.2$ & $35.6\pm0.2$ & $80.1\pm0.4$ & $92.1\pm0.3$ & $93.1\pm0.2$ & $96.6\pm0.1$ \\
& \checkmark & & $92.6\pm0.2$ & $41.8\pm0.8$ & $82.6\pm0.4$ & $93.1\pm0.6$ & $94.0\pm0.2$ & $96.6\pm0.3$ \\
& \checkmark & \checkmark & \textbf{92.8$\pm$0.2} & \textbf{42.4$\pm$0.7} & \textbf{84.0$\pm$0.4} & \textbf{93.2$\pm$0.8} & \textbf{94.1$\pm$0.2} & \textbf{96.7$\pm$0.4} \\
\bottomrule
\end{tabular}
\end{table}

\section{Limitations and Conclusion}
\label{limitations and conclustion}

\subsection{Limitations \& Future work}
\label{limitation}

Our approach, while robust in many scenarios, is not without its limitations. One notable constraint is the requirement of inputting two images during inference, which may not be feasible with certain scenarios where single-image processing is crucial. Despite this, our method still demonstrates a significant performance enhancement in most cases.

Moreover, while our framework achieves SOTA performance in the zero-shot inductive learning setting, it reveals a gap when compared to SOTA models trained under full-shot regimes and zero-shot transductive learning approaches. As \cref{tab: comparison with SOTA} shows, our method outperforms many existing models in unified AC and AL. However, it falls short of the benchmarks set by UniAD \cite{uniad} and MuSc \cite{musc}, particularly in scenarios where MuSc excels using visual features alone. This discrepancy suggests that there is substantial untapped potential for further exploration of visual reference features.

Additionally, our study offers novel insights into the application of the CLIP model for fine-grained anomaly detection: We demonstrate that joint visual and textual discrimination is a key contributor to enhancing fine-grained anomaly localization capabilities within the CLIP framework. Our findings also indicate that even when the visual reference images are anomalous, they can still serve as references for accurate anomaly scoring. These insights not only affirm the effectiveness of our proposed method but also open avenues for future research in refining visual-language models for more precise and versatile anomaly detection tasks.

\subsection{Conclusion}
\label{conclusion}

In this study, we introduced an innovative framework, the Dual-Image Enhanced CLIP for anomaly classification and localization, in the realm of zero-shot learning. Our approach leverages pairs of unlabelled images utilizes the pseudo anomaly in the TTA module, and demonstrates remarkable enhancement in performance, outperforming several SOTA methods. This advancement was achieved without the need for additional training, showcasing the framework's practicality and efficiency. Our findings also highlighted the untapped potential in combining textual features and visual references, suggesting room for further exploration in this domain.

\clearpage

%
%
\bibliographystyle{splncs04}
\bibliography{main}
\end{document}


\nolinenumbers

\title{Supplementary Material: Dual-Image Enhanced CLIP for Zero-Shot Anomaly Detection} 

\titlerunning{DICE for Anomaly Detection}

\author{First Author\inst{1}\orcidlink{0000-1111-2222-3333} \and
Second Author\inst{2,3}\orcidlink{1111-2222-3333-4444} \and
Third Author\inst{3}\orcidlink{2222--3333-4444-5555}}

\authorrunning{F.~Author et al.}

\institute{Princeton University, Princeton NJ 08544, USA \and
Springer Heidelberg, Tiergartenstr.~17, 69121 Heidelberg, Germany
\email{lncs@springer.com}\\
\url{http://www.springer.com/gp/computer-science/lncs} \and
ABC Institute, Rupert-Karls-University Heidelberg, Heidelberg, Germany\\
\email{\{abc,lncs\}@uni-heidelberg.de}}

\maketitle

\section{Dataset Preprocessing}

We adopted the OpenCLIP's \cite{openclip} outlined preprocessing methods for MVTecAD \cite{mvtec} and VisA \cite{visa}. The process commenced with the bilinear resizing of the images to a standard height dimension of $240$ pixels, coupled with a subsequent channel-wise normalization process. The VisA dataset posed a unique challenge due to its assortment of non-square images, which did not conform to the desired $(240, 240)$ dimension post-resizing. To address this discrepancy and ensure compatibility with the CLIP model's training dataset dimensions, we deployed the image tiling technique, which involved segmenting each image into two equal parts of $(240, 240)$. These segments were later merged back into a single image. Post-inference, the overlapping areas are averaged to maintain consistency in the final image representation.

\section{Prompt Templates}

We follows AnoCLIP \cite{anoclip} to produce prompts descriptions. It's composed of base templates, descriptive state words, and domain-aware prompts, denoted as following: ``[c]'' represents each class category; ``[s]'' denotes the state prompts; ``[d]'' is the domain-aware prompts.
By systematically substituting ``[s]'', ``[d]'', and ``[c]'' into the base templates, we generate a diverse array of prompts. These prompts effectively encompass both normal and anomalous scenarios within their respective domains.

\begin{multicols}{2}
\begin{itemize}
    \item \textbf{Base Templates}
    \begin{itemize}
        \item ``a [d] cropped photo of the [s]''
        \item ``a [d] cropped photo of a [s]''
        \item ``a [d] close-up photo of a [s]''
        \item ``a [d] close-up photo of the [s]''
        \item ``a bright [d] photo of a [s]''
        \item ``a bright [d] photo of the [s]''
        \item ``a dark [d] photo of the [s]''
        \item ``a dark [d] photo of a [s]''
        \item ``a jpeg corrupted [d] photo of a [s]''
        \item ``a jpeg corrupted [d] photo of the [s]''
        \item ``a blurry [d] photo of the [s]''
        \item ``a blurry [d] photo of a [s]''
        \item ``a [d] photo of a [s]''
        \item ``a [d] photo of the [s]''
        \item ``a [d] photo of a small [s]''
        \item ``a [d] photo of the small [s]''
        \item ``a [d] photo of a large [s]''
        \item ``a [d] photo of the large [s]''
        \item ``a [d] photo of the [s] for visual inspection''
        \item ``a [d] photo of a [s] for visual inspection''
        \item ``a [d] photo of the [s] for anomaly detection''
        \item ``a [d] photo of a [s] for anomaly detection''
    \end{itemize}

    \item \textbf{Descriptive State Words} \\
    normal states:
    \begin{itemize}
        \item s := ``normal [c]''
        \item s := ``unblemished [c]''
        \item s := ``flawless [c]''
        \item s := ``perfect [c]''
        \item s := ``[c] without flaw''
        \item s := ``[c] without damage''
        \item s := ``[c] without defect''
    \end{itemize}
    abnormal states:
    \begin{itemize}
        \item s := ``damaged [c]''
        \item s := ``abnormal [c]''
        \item s := ``imperfect [c]''
        \item s := ``blemished [c]''
        \item s := ``[c] with flaw''
        \item s := ``[c] with damage''
        \item s := ``[c] with defect''
    \end{itemize}
    \item \textbf{Domain Prompts}
    \begin{itemize}
        \item For all categories:
            \begin{itemize}
                \item d := ``industrial''
            \end{itemize}
        \item For surface categories (carpet, leather, grid, tile, wood):
            \begin{itemize}
                \item d := ``textural''
                \item d := ``surface''
            \end{itemize}
        \item For all other categories:
            \begin{itemize}
                \item d := ``manufacturing''
            \end{itemize}
    \end{itemize}

\end{itemize}
\end{multicols}

\section{Ablation Study on Hyperparameters}
We conducted a comprehensive performance comparison across various settings of hyperparameters $\lambda_1$ through $\lambda_5$. These experiments were executed using $6$ different random seeds, and we report the results as mean values with standard deviations to provide a clear understanding of variability and reliability.

\begin{table}[]
\caption{Comparative study of the zero-shot anomaly localization (AL) performance on MVTecAD with various $\lambda_1$ and $\lambda_2$ settings. Bold values indicate the best results.}
\label{tab: your_label}
\centering
\begin{tabular}{@{}p{1cm}p{1cm}p{2cm}p{2cm}p{1.5cm}@{}}
\toprule
$\lambda_1$ & $\lambda_2$ & AUROC & F1Max & PRO \\
\midrule
1 & 0.5 & $92.5 \pm 0.2$ & $41.7 \pm 0.8$ & $82.2 \pm 0.3$ \\
1 & 1 & $92.7 \pm 0.2$ & \textbf{42.4$\pm$0.7} & $83.4 \pm 0.4$ \\
1 & 1.5 & \textbf{92.8$\pm$0.2} & \textbf{42.4$\pm$0.7} & $84.0 \pm 0.4$ \\
2 & 2 & $92.7 \pm 0.2$ & $42.3 \pm 0.6$ & \textbf{84.1$\pm$0.3} \\
\bottomrule
\end{tabular}
\end{table}

\begin{table}[]
\caption{Comparative study of the zero-shot anomaly classification (AC) performance on MVTecAD with various $\lambda_3$, $\lambda_4$, and $\lambda_5$ settings. Bold values indicate the best results.}
\label{tab: new_label}
\centering
\begin{tabular}{@{}p{1cm}p{1cm}p{1cm}p{2cm}p{2cm}p{1.5cm}@{}}
\toprule
$\lambda_3$ & $\lambda_4$ & $\lambda_5$ & AUROC & F1Max & AP \\
\midrule
1 & 1 & 1 & \textbf{93.2$\pm$0.8} & \textbf{94.1$\pm$0.2} & $96.7 \pm 0.4$ \\
1 & 1 & 2 & $93.1 \pm 0.8$ & $94.0 \pm 0.3$ & \textbf{96.8$\pm$0.4} \\
1 & 2 & 1 & $91.9 \pm 0.8$ & $93.2 \pm 0.2$ & $94.8 \pm 0.4$ \\
2 & 1 & 1 & $93.1 \pm 0.6$ & $94.0 \pm 0.1$ & $96.7 \pm 0.3$ \\
\bottomrule
\end{tabular}
\end{table}
\section{Visualizations}
\begin{figure}
    \centering
    \includegraphics[width=\textwidth]{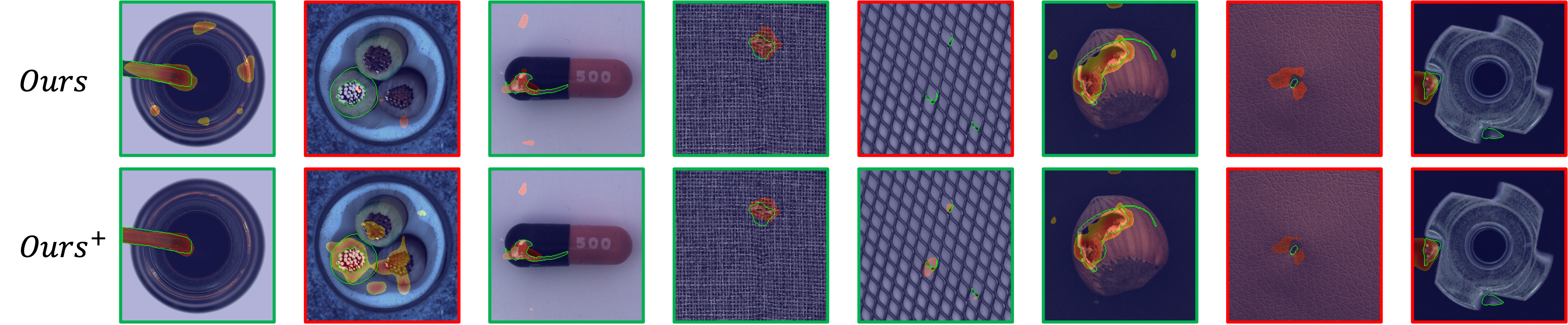}
    \caption{Visualization of prediction examples from ``bottle'', ``cable'', ``capsule'', ``carpet'', ``grid'', ``hazelnut'', ``leather'', and ``metalnut'' categories. Green line in the images denotes the ground truth of the anomaly. The success and failure cases are bordered with red and green, respectively.}
    \label{fig:supp_vis1}
\end{figure}

\begin{figure}
    \centering
    \includegraphics[width=0.9\textwidth]{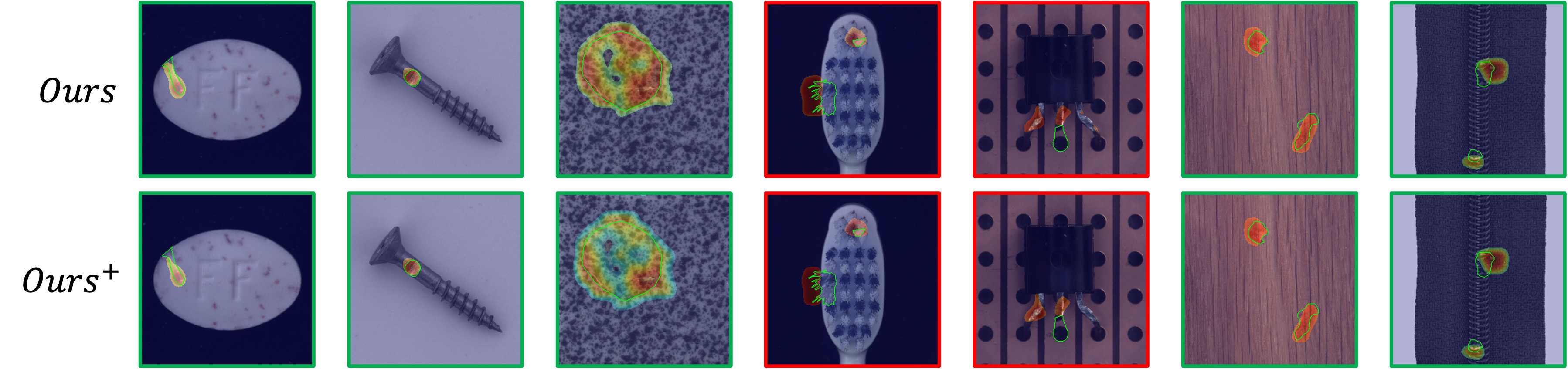}
    \caption{Visualization of prediction examples from ``pill'', ``screw'', ``toothbrush'', ``transistor'', and ``zipper'' categories. Green line in the images denotes the ground truth of the anomaly. The success and failure cases are bordered with red and green, respectively.}
    \label{fig:supp_vis2}
\end{figure}

\begin{figure}
    \centering
    \includegraphics[width=\textwidth]{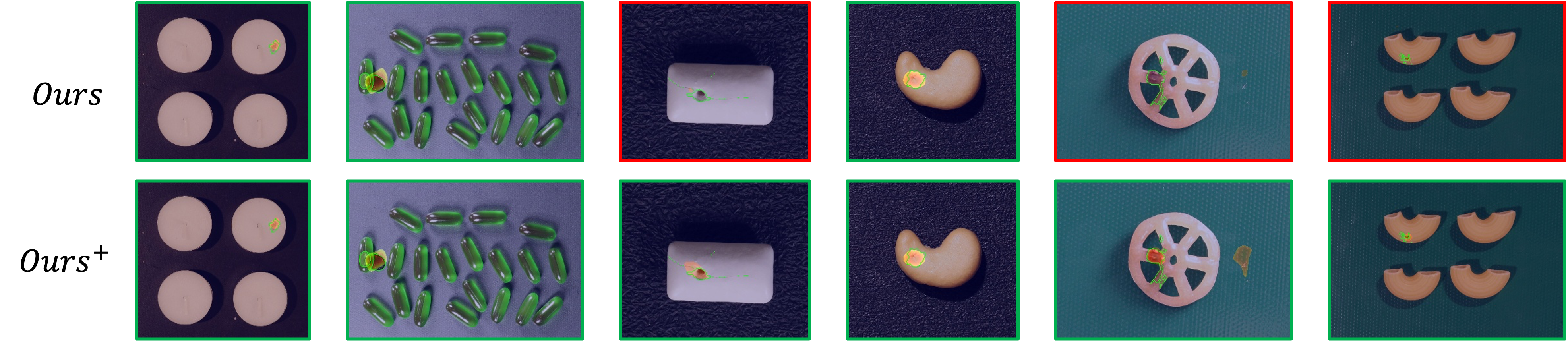}
    \caption{Visualization of prediction examples from ``candle'', ``capsules'', ``chewimg gum'', ``cashew'', ``fryum'', and ``macaroni1'' categories. Green line in the images denotes the ground truth of the anomaly. The success and failure cases are bordered with red and green, respectively.}
    \label{fig:supp_vis3}
\end{figure}

\begin{figure}
    \centering
    \includegraphics[width=\textwidth]{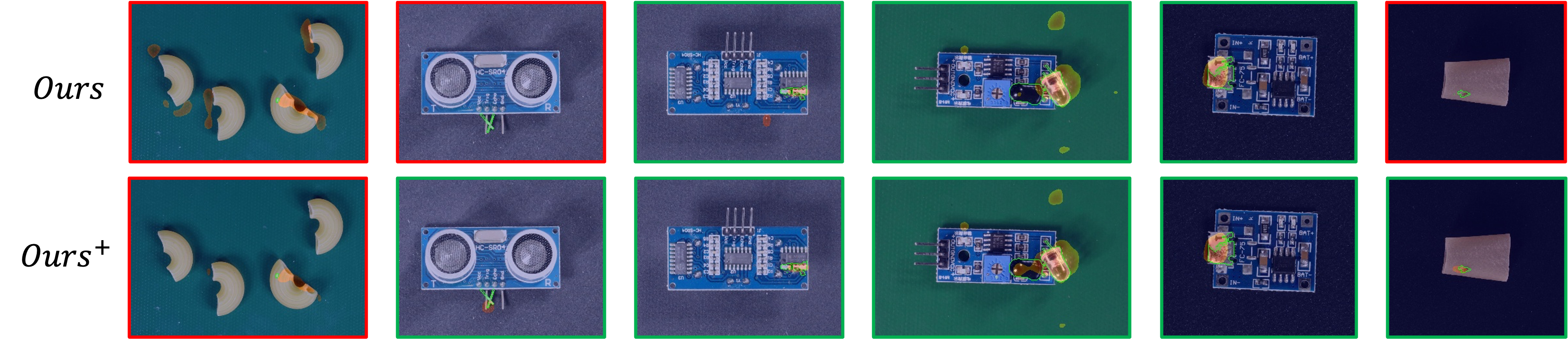}
    \caption{Visualization of prediction examples from ``macaroni2'', ``pcb1'', ``pcb2'', ``pcb3'', ``pcb4'', and ``pipe fryum'' categories. Green line in the images denotes the ground truth of the anomaly. The success and failure cases are bordered with red and green, respectively.}
    \label{fig:supp_vis4}
\end{figure}

We visualize the zero-shot anomaly detection results on MVTecAD in \cref{fig:supp_vis1} and \cref{fig:supp_vis2}
and VisA in \cref{fig:supp_vis3} and \cref{fig:supp_vis4}. Notably, the segmentation threshold is selected based on the max value of the F1 score. Also, our TTA enhanced method which is denoted as $ours^{+}$, yields more accurate and comprehensive results, even in difficult scenarios and failure cases.

\clearpage

%
%
\bibliographystyle{splncs04}
\bibliography{main}